# Employing Socially Assistive Robots in Elderly Care (longer version)

Daniel Macis

University of Turin, ICT School, daniel.macis.1992@gmail.com

Sara Perilli

University of Turin, ICT School, sara.perilli.1997@gmail.com

Cristina Gena

University of Turin, Department of Computer Science, cristina.gena@unito.it

Recently, it has been considering robotics to face world population aging. According to the WHO, in 2050 there will be about 2.1 billion people over 60 years old worldwide causing a persistent growing need of assistance and a shortage of manpower for delivering congruous assistance. Therefore, seniors' QoL is continuously threatened. Socially Assistive Robotics proposes itself as a solution. To improve SARs acceptability, it is necessary to tailor the system's characteristics with respect to the target needs and issues through the analysis of previous and current studies in the HRI field. Through the examination of the state of the art of social robotics in elderly care, past case studies and paper research about SARs' efficiency, it has been proposed two potential solution examples for two different scenarios, applying two different SARs: Pepper and Nao robots.

**CCS CONCEPTS** • Human-centered computing • Human computer interaction (HCI) • Interactive systems and tools

**Keywords:** HRI, Care Robots, Socially Assistive Robotics, Elderly Care

## 1 Introduction

Recently, one of the biggest problems we are facing is the world population ageing which causes an increase of people in need of help with their physical and mental health coping, for instance, with Dementia. According to the World Health Organization (WHO)1, in 2050 about 2.1 billion people worldwide will be over 60 years old meaning an increase in assistance requests, causing for certain a further worsening of the current shortage of caregivers [1]. Nowadays cutting-edge technology, as robotics itself, may represent a proper solution to the rising problem integrating socially assistive robots (SARs) in everyday life of elders and caregivers, to respectively improve their Quality of life (QOL) and working conditions. SARs are embodied systems meant to assist the user mentally and physically through social interaction. Fong [2] identifies the key features that SARs need to be defined as such: embodiment, intentionality, socially situated learning, user modelling, dialogue capabilities, human-oriented perception, emotionality and personality. In the last few years, the perception of robots as industrial devices has changed, granting them application in domestic and social purposes. A consistent branch of HRI research focused on this topic, striving to demonstrate the effectiveness of SARs in elderly care and scoping the acceptability of the robotic systems [3]. Nevertheless, realising an efficient and effective system without a deep investigation of the target requirements and issues may turn out to be complicated. Therefore, it is fundamental to inquire about caregivers and elderly's set of problems to diminish the churn of the assistive system and prototyping SARs able to fulfil recipients' real needs.

---

1 https://www.who.int/news-room/fact-sheets/detail/ageing-and-health#:~:text=At%20this%20time%20the%20share,2050%20to%20reach%20426%20million.

This paper has been organized as follows: Section 2 reviews the SARs background in elderly care, Section 3 presents related work with implementation examples of SARs, Section 4 presents some possible scenario, while Section 5 presents some possible solution, and finally Section 6 concludes the paper.

## 2   Sars Background in Elderly Care

Investigating a potential target needs and issues results crucial not only to realise a useful system but also to comprehend how to steer the prototyping of the latter to the user, focal point to focus on to make the final user to perceive the system as useful and helpful, affecting the acceptability degree of such technology and its intention to use.

## 3   Acceptability

The acceptability of a technological system may be influenced by features deriving from the user's personal traits. During the analysis it is compelling not to stereotype or to trust false impressions which may lead to incorrect inferences. It is not uncommon to take for granted that age is directly related with the acceptability of advanced technological systems, as SARs themselves are. Contrary to popular belief, research highlights that it is not the user's age influencing the system's perception and intention to use [4]. The educational level and experience with technology directly affect the latter [5], as users keener on technology and with higher educational levels resulted more open minded and trustful towards robotics in everyday contexts [4], thus it is worthwhile to involve users since the primary school in activities as coding and educational robotics [50]. Moreover, it is likewise to consider the user's cultural background. It is common knowledge and stated that in eastern countries, such as Japan, SARs are broadly employed in many aspects of everyday life and therefore well appreciated. On the contrary, in Europe people think that robotics must be employed mainly in perilous contexts otherwise in grunt work. Consequently, there is still a lot of mistrust regarding employing SARs if employed in elderly care [6][7] [8] [9]. Focusing on seniors, research highlighted that the necessity to preserve their independence and to have someone to lean on could influence SARs acceptability, suggesting that lonelier seniors are also more prone to accept them [10] [11] [12]. In addition, perceived usefulness, perceived ease of use, social capabilities, proper design, privacy, ethical regards and uncanny valley avoidance are key points in acceptability as well. Quite the contrary, usually workers in the healthcare field have different points of view about this topic: often they are frightened SARs could replace them in their job, sometimes they don't think they are able to use the system otherwise SARs seem like a burden in ordinary workload. Furthermore, they are afraid that seniors could be accidentally hit by SARs, or they could lead seniors to be even more isolated than usual. However, those fears are led by the lack of knowledge about the subject of study provoking a wrong perception of socially assistive robots and their role [13] [14].

### 3.1   Problems and needs

As already mentioned, it has been employed a user-centred approach to analyse seniors' insight [15]. In the first place, everyday problems were considered. Several research highlighted senior's major issues and needs through some surveys conducted by asking seniors but also professional caregivers and non-professional ones. Actually, the latter have a key role in the investigation being stakeholders and part of the senior's everyday life. Consequently, they allow a better comprehension of everyday routine and issues in elderly care. Major needs and issues emerged are shown below:
- Loneliness and feeling of rejection are two of the major issues which cause depression and bad mood in seniors: it is necessary to neutralize loneliness and isolation and to support the seniors emotionally and psychologically [6] [16] [14] [12]. In this field, the SARs' social characteristics appear as a perfect answer to the problem. Actually, one of the necessities is represented by the desire of conversing with someone who listens to them [12] [16], to play with and to talk about repetitive topics, like weather forecast and similar [16] [17]. From another point of view, offering easy access to communication technologies to seniors could also bring advantages [14] [17]. Through a user-centred approach, which means an easy mode of operation, it is possible to bring together technology and elderly, making it possible for seniors to get in touch with friends and family.
- Physical decay and mobility problems: sometimes seniors are frightened by moving alone because of the falling risk due to the lack of balance [14]. That brings to light the need for closer supervision which sometimes may not be possible for caregivers to do, meaning a 24/7 oversight of the elderly to prevent falls and to immediately provide first aid.
- Mental health and memory decline are other relevant concerns for seniors [17]. In these contexts, it is necessary to provide them with devices to help them remember important events, their daily pharmacological routine, mealtimes and to drink [14] [17]. Mental health could be preserved or kept trained with specific activities conceived for seniors [18, 16]. In addition, activities like these may help patients to spend their time more pleasantly.

Some other issues and needs are lightened up only by caregivers: caregivers express the desire of having support with grunt work, such as going along with seniors from one place to another. These activities occupy caregivers' time, preventing them from spending quality time with the assisted ones [19]. Moreover, it seems essential to overwatch not only the patients' movements, but also their vital signs as well [20] [21].

## 3.2 Main benefits of SARs usage

Reasons to implement SARs in elderly care are solid and full of benefits, in many cases fully or partially restoring the autonomy of the elderly. Some of the major benefits are exposed below:
- Integrating telemedicine to allow a remote treatment of patients, especially in nursing homes [14].
- Helping caregivers in their job and in monitoring seniors 24/7. It could help lighten the ordinary load of work caregivers have to face every day, also lowering their degree of stress [6].
- To have a reminder that helps seniors who live alone to remember events and medicament delivery [6]. This could also help caregivers which sometimes can forget essential matters due to stress, to minimise error risks and forgetfulness.
- Implying SARs in grunt work permits caregivers to spend more quality time with the elderly, otherwise spent in low-skilled tasks [19].
- Research proved that SARs' social abilities influence seniors' QoL in a positive manner [5]. As a matter of fact, SARs' capabilities positively affect anxiety levels, stress and depression. Moreover, they succeed in reducing loneliness, improving physical and mental health [5] [19]. In addition, SARs appear useful to improve rehabilitation therapies efficacy more than nowadays technological devices such as computers, smartphones or fitness trackers can do [20].
- SARs potential dwells also in helping emotionally the patient thanks to their empathic skills. These capabilities may also be employed with cognitively impaired seniors who tend to isolate themselves due to discomfort, confusion or moodiness [6] [21] [22].
- Thanks to SARs' social and empathic skills, socialisation among seniors themselves and between seniors and caregivers is encouraged [23]. Taking advantage of social skills could be useful to make integration of new residents in nursing homes simpler, helping seniors to make new friends [6].

- SARs can be applied in some specific therapies already used in elderly care, such as pet therapy and doll therapy. Using SARs make seniors more engaged rather than with classic therapy, especially in pet therapy because of some seniors' fear towards pets [21] [12] [24] [25] [26] [27] [28] [29].
- Using socially assistive robots instead of humans for close approaches with seniors when not unavoidable can minimize the transmissibility risk of COVID-19 during the pandemic emergency [30].

## 3.3 Main entry barriers

These kinds of systems proved to be promising but nowadays they are used only by few research institutes or nursing homes because of their many entry barriers. First of all, it is right and proper to express the fallibility of robotic systems which means that the robot is not completely safe as desired, due to potential robot's failures and jams in its interaction or due to sensors' liability to error. Problems like these could stoke some existing preconceptions and prejudices [9].

In particular, caregivers reported the fear for elderly safety when they are asked to express an opinion about the employment of social robots in elderly care [13] [14].

The fear is also stoked by the few experiences with technological systems and issues in general. At the same time, it is necessary to also discuss the opposite problem. It is also incorrect blindly relying on the systems not considering potential failures, activating a bias which depicts the systems completely perfect and free from errors. Lastly, it is essential to mention one of the biggest entry barriers: the high price of the robot that impedes SARs adoption and spread where they could be needed and useful.

## 4 State of the art of SARs in healthcare

Research has put in place the realisation of various types of SARs, devised for different contexts and targets, primarily aiming to support caregivers and not to supersede them [31]. We can encounter different employments of SARs in healthcare: impersonating assistants, supporting therapies, used to improve social and communicative life, becoming everyday companion robots thanks to their empathic capability, a fundamental and irreplaceable SARs' characteristic [14]. It is possible to distinguish different types of SARs employed in healthcare, such as pet-like robots and human-like robots.

### 4.1 Pet-like robots

Pet-like robots are all those robots whose appearance is like real pets and for this reason they are mostly employed in pet-therapy contexts with fragile patients in need of affection. Some examples are Paro, Aibo and iCat, useful in reducing stress and anxiety in seniors with dementia diagnosis. These kinds of robots are used thanks to their abilities in helping seniors' social lives, reducing feelings of loneliness [24] and improving verbal and physical capabilities [46]. As an example: Paro is a puppy seal-like robot, it is dressed in fur and covered with touch detectors. As already mentioned, Paro is employed in nursing homes to assist seniors with a dementia diagnosis because of its abilities in reducing stress and anxiety levels and in bringing back to normal blood pressure levels. Actually, during research tests, Paro improved relationships among seniors and reduced analgesic usage, unburdening at the same time caregivers' load of work. [12] [24] [25] [26] [27] [28] [29].

### 4.2 Human-like robots

Human-like SARs are morphologically like humans to improve acceptability and intention to use, being employed in collective or domestic environments with assistential duties. They implement speech synthesis and communicative capabilities, not just verbal skills, but also such as facial expressivity, gestures, body posture communication, face tracking and gaze direction. It is possible to interact with them also with haptic devices and through displays. Often endowed with affecting computing capabilities to cognitivize and empathise with patients, they can also learn and mimicry thanks to machine learning programs [31].

### 4.2.1 Pearl

We can consider Pearl as one of the first SARs for elderly care, initially devised for domestic use and subsequently employed in RCFEs[2] and specialised healthcare centres, in which it is necessary a continuous and specific treatment for elderly who often suffer from chronic disabling diseases. This social robot has been designed to assist seniors in their daily routine through a daily events scheduling, reminding them to drink and to take necessary drugs, helping them reach the toilet and with orientation, directing the elderly where they need to go. Giving directions to seniors is a widely appreciated skill by caregivers, as they usually spend a lot of time guiding residents through the runners of the facilities. Lastly, Pearl can speak and communicate through facial expressions and physical movements allowed by its DoFs, integrating the interaction with a touch-based display [32] [33].

### 4.2.2 Ryan

Ryan is an individual use devised SAR designed to assist elderly affected by early onset BPSD[3]. Ryan implements an oval-shaped display in which the personalized face of the robot is projected for expressive purposes. It's competent in face tracking and presents a touch screen placed in the torso to interact with the SAR and to play music and videos, show digital pics and meant for visual cognitive and reminiscent exercises. Cognitive games are non-pharmacological therapies effective in slowing down the mental decay of the patients, usually exploited in BPSD treatment therapy, improving mood, wellness and stimulating seniors' mental skills [34]. The system is personalized in musical tastes, videos, photos and conversational topics modelled following the patient's characteristics, daily collected through the log analysis of the interaction between the resident and Ryan. This SAR also presents practical skills, as daily reminders are. It is considered astonishing the huge number of interactions with Ryan, also taking in consideration the presence of various social activities in the care facility, so it's not surprising that the SAR's test disposal has been greeted by elderly with sadness and melancholy, who were satisfied and involved in the robot's use [34].

### 4.2.3 Tangy

Bingo is one of the most appreciated recreational activities in nursing homes because it is a cognitive training but also a good social activity for seniors. Tangy is a robot developed with the aim to assist healthcare workers during bingo sessions. Actually, it can play announcing drawn numbers with its voice, helping itself with its tablet, and assisting users who call for help. The game starts getting Tangy in front of the gamer's table. Each gamer owns a bingo card and a push button that they can use when they need help to report an event. Pushing the button, the user lifts a silver infrared reflective triangle attached to the table so that Tangy could identify and move towards the user to help him. Tangy was tested with over 60 fluent English users who don't have hearing problems and dementia diagnosis. Results of the test reveal a game success rate of 100% due also to Tangy's ability to assist users, despite users in need of help identifying problems. However, that may be, the interaction appeared quite fluent and genuine, engaging and stimulating socialisation [35].

### 4.2.4 Betty

Betty is a small sized social robot whose appearance is like that of a child. Betty is capable of vocalisations, face recognition, face tracking, face recording, communicating through facial expressions and gestures, emotion recognition and can recognise environmental sounds. Betty can also dance and sing. Its interaction is possible through speech, touch and a touch-based tablet provided with a graphical interface. Research suggests that the embodiment of a virtual agent in a human-like shape, concurrently with multimodal interactions encourages the acceptability, the usability and intention to use of the robotics system, with a relatively high level of engagement. The robot's most appreciated skills have been dancing and singing, followed by weather forecast and news reading.

---

[2] Residential Care Facilities for the Elderly
[3] Behavioral and Psychological Symptoms of Dementia

It is easily inferable how non-functional skills are useful in social therapy, in particular with patients affected by BPSD. Lastly Betty has also permitted the caregivers to relax and to have more time to deliver other kinds of assistance to the seniors [36].

### 4.2.5 Pepper

One of the most appreciated SARs is Pepper, a 1.20 m tall teenager-like humanoid robot. Thanks to its vocal ability and its gesturalism it can simulate a human-like dialog helped with the support of the tablet on its chest. Lots of research has proved its efficacy in elderly care settings. An example is German research conducted in a local nursing home [46]. Here, despite the expected initial mistrust of seniors, the latter revealed interest and acceptability towards Pepper, especially when they played some activities with the purpose for physical help, cognitive training and socialisation purposes. Some activities, such as "guess the song" game works better than others because of music special effects such as recollection of memories, then allowing a genuine and friendly interaction [47]. Often, users consider Pepper as a kid, for this reason tending to forgive its errors and weaknesses, such as limits in dialogue abilities and interactions failures. Alternatively, Pepper4Elderly project [30] tries to imagine Pepper in a similar background, playing the solution to some classic problems, like for example monitoring seniors' health and mood while considering new issues related to pandemics too. Main purposes are to try to restrain contagions by monitoring all seniors' physiological parameters, remind them of safety measures and rules and at the same time guarantee them to stay in touch with families. At least, Pepper should provide entertainment and encourage body and mind wellness, for a continuous analysis of cognitive decline, also making use of specific tests.

### 4.2.6 NAO

NAO is a 57 cm tall humanoid robot, whose aspect is similar to a baby for acceptability reasons. It is capable of speech recognition, speech synthesis and emotion recognition. NAO has been employed with individuals with MCI[4] designing specific activities for the patients [48] [49]. Moreover, A NAO's variant called Zora has been the protagonist of an important long-term study involving 245 residents of some Dutch RFCEs. The main goal has been scoping the SAR's application procedure in RFCEs, periodically recording the caregivers and users' experiences. Zora can socialise and guide the elderly through specific activities, such as cognitive games for BPSD affected patients. From the over 150 field game sessions observations emerge how the mood of participants improves during the playful activity and returning to basic levels once finished. Also, engagement follows a similar tendency. Zora's dancing and playful skills, such as the bingo game, have been considered amusing and engaging by residents, as well as musical reproduction for reminiscent purposes. Caregivers and managers too have proved themselves favourable to use SARs in RFCEs. Even though the experimentation met several problems, such as the limited robot's battery capacity, problems with the SAR's speech recognition and speech synthesis, as well as robot's software updating problems, Zora proved to be able to stimulate the spontaneous participation of patients to activities, also being useful to calm states of anxiety and agitation in restless seniors, also maintaining its attractiveness over time [37].

### 4.2.7 ARI

In the last years we witnessed the development of the humanoid robot Ari, designed with special regard towards the user-centered design principles, distinguishing itself from competitors for its particularly anthropomorphic features, especially in height and speech. Using natural language processing, face and object recognition, concurrently with advanced machine learning programs, Ari is capable of imitating the user's behaviour and thanks to facial expressions, tonality of voice, body postures and gestures. Like many SARs, Ari too has been devised to guide users in hospital facilities, to hold receptionist roles or to entertain people. In domestic or collective environments, such as RFCEs, Ari can monitor patients' health, be a playful therapeutic company or a playfellow through the tablet it implements in the chest, meant also for telemedicine purposes [24].

---

[4] Mild Cognitive Impairment

# 5 SCENARIOS EXAMPLES

It is necessary to outline the users' general profiles and Personas, personalising the interaction and the affective computing of the robotic systems on the real needs of the target, for details see [51]. It is also necessary to hypothesize a congruent background and scenario to realise an interaction tailored to the target. Two user's profiles have been defined: the first one proposes a solution to deliver SAR's home care, while the second one takes into account inserting a SAR into a nursing home.

## 5.1 Scenario 1: Home care

In the first hypothesized scenario our Persona is called Armando. He is 78 years old and was left a widower. He is self-sufficient, living alone in Turin. He has a 45 year old son, who lives in Milan. Armando has a sedentary life, spending a lot of time alone at home. Fearing for his safety, Armando feels himself vulnerable to strangers who may sneak into his abode. Being elderly, his memory is no longer that strong, needing help to remember appointments and important daily events, such as taking his medicines, drinking or his medical visits. Formerly it was his wife who reminded him of relevant events. Armando would also like to stay informed on the latest news, on weather forecasts and on topics of interest. He is no longer interested in TV programs as in the past, feeling them distant from his interests.

Since he became a widower, Armando has significantly reduced his social contacts, retreating into himself and suffering from solitude. He needs to be stimulated to keep contact with the loved ones. However, Armando loves to converse and to be listened to by someone, failing in satisfying this kind of need during the day. Even leisure opportunities have become rare, and Armando would like to have more fun. His mental and psychological health are affected by loneliness. Armando is in need of reassurance and solace due to his anxiety and panic attacks, a phenomenon from which he suffers since he became a widower. He would also like to feel safe in case of illness.

His son Giacomo tries to take care of him on the weekend, when he is not at work, visiting him with her 10 years old daughter Cecilia, though he cannot spend much time with him. When Giacomo is with his father, he realises that Armando loves spending time in company of him and Cecilia. Giacomo would like someone to keep Armando company, helping him with his necessities. Unfortunately, Armando does not tolerate strangers well in his house, as for instance a human caregiver could be.

## 5.2 Scenario 2: Nursing home assistance

For the purpose of implementing a SAR in a collective context, it is also possible to imagine a Persona who lives in a nursing home with different needs and issues. For example, Ms. Tina is an 81-year-old woman. She lived in a very big house, in the countryside far from her sons and grandchildren, too big and difficult to manage for a widow. Pandemics and lockdown made her depressed because her social activities were drastically reduced. Also, aging Tina began to suffer from some health problems, such as diabetes. Her sons fear that she would let her health go but they cannot take care of her because of their job and family responsibilities. Moreover, they preferred that an expert would take care of that. Therefore, Tina and her sons thought that living in a nursing home was the best option to have the health conditions monitored but also with the hope that her social activity could be better than before.

Tina's typical day is simple and monotonous: she spends most of her time alone watching tv and reading magazines. Sometimes she has some health exams for her check-up routine, but she does not attend any activities to train body and mind or to make new friends. As a matter of fact, Tina's social life has not changed since she has been living in the nursing home: she does not bond with peers nor with caregivers, with whom she speaks only if necessary and for this reason she feels abandoned and it seems she is a burden for them. Actually, depression and loneliness are still part of Tina's life. Tina needs some incentives to make her day dynamic and not make her feel lonely. It is necessary to create opportunities to develop passion, to train her body and brain and to make new friends in the nursing home to have someone to speak, play and have fun with and to spend her last years peacefully.

In the end she also would like to meet her sons and grandchildren more often but with pandemics all the meeting affairs are changed and visits are strongly limited so that she can only call her relatives just like she did when she was at home.

# 6 Solution Proposals

## 6.1 First scenario's solution proposal

### 6.1.1 NAO robot Persona

Thanks to its small size and sensors, the preferable robot for domestic use is NAO. Its primary target are older adults living in loneliness conditions within home or living with a professional or non-professional caregiver. Once defined the user's Persona, it is necessary to define a robot's Persona (for details see [45]) too in order to personalise NAO's AI and to make the interaction more believable and empathic, also keeping the engagement levels at high. This will influence the SAR's behaviour in the various situations, the impersonated role, its tone of voice and the utilised speech register.

In this use context, NAO holds the role of a 12 years old personal assistant who lives in Turin. For engagement purposes he pretends to be a boy scout, whose job is to assist the elderly. Holding this long-term duty, NAO will try to establish a grandson-grandfather similar relationship. NAO uses a colloquial register, always trying to be understandable. It is however aware of useful notions, being always in search of knowledge and curious by nature. He has a friendly ToV and contextually enthusiast for the happenings and interactions with people. He has a simple sense of humor, knows simple jokes and puns and also loves guessing games. His epistemic confidence in the user is relatively high, trusting people's collaboration in dialogues and interactions. NAO is very sociable and careful with the assisted one, doing everything possible to support, keep in good health, entertain and make feel better the elderly, as a lovely nephew would do. NAO is proactive in entertainment, being servile when necessary. He keeps the situation under control when not busy in helping activities. He listens and satisfies the user's requests, trying to improve based on experience in interactions.

In case of foul language NAO tries to dampen the situation trying to make the elderly desist from repeating the offenses, having the social robot a pacific and sociable soul.

### 6.1.2 NAO's framework solution

For the use of the robot NAO as a companion robot it has been developed a framework of interactions and activities to assist the elderly based on its needs. Within the domestic context, the SAR will offer assistance to the user in non-physical tasks, acting if necessary as an entertainment robot and improving senior's QoL, allowing him to keep contact with the society and representing a bridge to strengthen human-to-human social interactions.

NAO could also remind the elderly the scheduled medical visits, which need to be previously verbally communicated to the SAR by the user or possibly by the caregiver, in order to insert the event into memory and to schedule it, allowing to remind memorandum and dates deemed relevant by the senior. The social robot will also be able to remind the user which medicines take and their relative dosage, announcing it at the scheduled time, thus checking if the elderly actually took its medications. Similarly, the SAR could periodically remind the elderly to drink and eat, checking that the action is performed.

At each meal, NAO could also recommend a list of dishes to the elderly suggested on the food diet to follow, memorising the expressed preferences and previous requests. If requested, the SAR could also accompany the senior to the kitchen to assist him preparing the meals.

For informative purposes, NAO could update the senior on the latest news, on weather forecasts and answering general knowledge questions posed by the user.

NAO could offer help in case of necessity, calling people indicated as references and, in the second place, emergency services. In synergy with a wearable device, NAO could also be able to detect elderly's falls, autonomously warning of the event a pre-registered contact and asking the user about his current state of health, if necessary calling the emergency services. Using a wearable device, it is also possible to constantly monitor the senior's vital signs. Also, the social robot could monitorate and localise the user through the use of the available cameras.

With the SAR it is possible to schedule simple physical activities for the elderly to keep him healthy.

Using affective computing, NAO will be able to monitor user's emotions, mood, detect states of agitation and confusion, calling for help when requested or necessary. The SAR will thus reassure the senior in case of negative emotional states.

A specific activity for the management of negative emotional states, based on mindfulness, was also designed and developed, useful for relieving symptoms of physical and psychological discomfort.

NAO will also be capable of improving the user's self-confidence, gratifying him during the physical exercise or upon reaching the daily goals. Always with the aim of improving the emotional and emotive elderly's health, the SAR could solicit the user's brain functions with several cognitive games.

To counteract the social isolation, the system could stimulate the senior in keeping contact with the family and society through diversified methods, for instance suggesting him to call a relative in case of boredom.

NAO is competent in various entertaining skills, such as reading books, playing music, telling stories, dancing or singing with the user. The framework of activities also comprehends some guessing games, quizzes, and other kinds of games to play with the senior or more players, representing the robot a leisure mate.

### 6.1.3 Mindfulness application

The mindfulness application has been devised for the relaxation of the user, to control its states of anger and to hold in check situations in which the senior could experience anxiety or agitation. Mindfulness proves its effectiveness against the effects of depression and improving QoL [39], being employed to relieve symptoms of chronic physical and psychological discomfort, as anxiety and physical pain are [40].

Developing and implementing the mindfulness activity in SAR's capabilities offers the possibility to use it in a scheduled way, always having the activity available. The application is inspired by the mental relaxation exercises of the clinical psychologist Laura Pirotta [41], also taking inspiration from the structures of the mindfulness applications Meditopia [42] and Serenity [43].

The activity has been developed on the proprietary platform Choregraphe [38] and it is also usable by users without past experience in guided meditation, from the very moment that the SAR will guide them through the activity.

The interaction develops itself on several levels, being partially controllable by the user, mostly as regards its interruption. Several variants of the robot's utterances have been developed, comprehending numerous animations related to gesture, in order to make the dialogue more engaging. The application also presents some fallback functions, such as the management of cases in which the user expresses no utterances, managing repetition requests and comprehending functions related to the missed recognition of the concept expressed by the user. In certain situations, the system uses emotion recognition functionality, through the use of a database specifically created and implemented on Choregraphe, permitting the recognition of 12 different moods and emotions relying on sentiment analysis.

Preliminarily, the robot should contain into its knowledge the user's personal information, in particular his name, or alternatively being able to recognise the elderly, see for instance [51]. Calling the user by name will make the interaction more engaging.

When the application starts, NAO will greet the user ascertaining his health and emotive state. In case of recognition of negative states, the SAR will proceed asking the user the motivation about a certain state of mind, empathising consequently and, if deemed appropriate, offering a remedy through the mindfulness activity. In case of pain or physical discomfort, the social robot will ask the user whether to call for external help, sending a notification and calling an emergency contact, recorded in advance in the memory of the robot. In the latter eventuality, NAO will proceed towards closure, remaining available to the user.

Having the emotional and health state ascertained, the SAR asks the user whether proceeding with the mindfulness or not, asking him if he knows how it works and explaining the operation of the activity if requested. The SAR recommends the user to sit in front of him, waiting to start the activity on command of the user.

At the request to start the mindfulness, the robot will recommend the user to decrease the intensity of the lighting to promote relaxation, thus reproducing ambient music. NAO will then proceed with the first activity phase, telling the user the actions to execute, spaced 8 seconds apart and accompanied by gesture. SAR's utterances are extracted from different sets and randomly played at each move. After the first phase, the social robot will check the correct course of the interaction, interrupting it if requested or if the user should experience physical pain or strong discomfort. NAO would then ask the user whether to call for help, managing the situation and moving towards closure. If no problem occurs, the SAR will initiate the second phase of mindfulness, similar to the first stage.

The activity consists of 3 stages and, except the final phase, the social robot will ascertain the correct performance of the interaction at the end of each stage. The number of cycles can be extended to any number. Once the activity is finished, NAO will ask the user for feedback on the latter, which preference can be expressed on a 1 to 5 Likert scale, reacting consequently to the expressed score and recording the result in a log, to allow the developer further improvement.

In the end, the SAR will interrupt the music reproduction and will move towards closing, farewelling from the user and remaining at his disposal for any request.

## 6.2 Second scenario's solution proposal

### 6.2.1 Pepper robot Persona

A profile with that kind of necessity needs emotional support, social and amusement occasions but also a deeper attention to physical and mental health. For this reason, the chosen robot is Pepper, a SAR capable of social activities, able to stimulate socialisation among seniors and also capable of simulating empathy towards users. Pepper's job in the nursing home is a young aspiring therapist - social worker in public health service. Its assignments are about helping caregivers in better managing daily routine and taking their place in repetitive tasks and, on the other hand, Pepper should provide to seniors' entertainment, being their playmate, speaking with and encouraging them in physical and brain training that the robot manages. Pepper has a colloquial Tone of Voice because of the personality it has been thought of for its Persona: a curious young man with a cheerful and extrovert temperament. A good mood spirit identifies Pepper, a mood that it tries to transmit every day to seniors, supporting and motivating them. Pepper is friendly and open minded to new experiences encouraging others to do the same. Pepper also likes feeling helpful and for this reason it is willing to do something for others but without being intrusive with users. In the end, Pepper loves to accomplish its tasks precisely and with accuracy but always empathically.

### 6.2.2 Pepper's framework solution

To achieve this goal, users and Pepper can interact with each other in an orally way, which is the simplest form of interaction for users, but this is also possible to do through Pepper's tablet located on its chest. Pepper's tablet is useful in providing more visual-based and touch-based affordances that permits to carry on a better interaction, in particular for senior users which are not accustomed to technologies. Moreover, to simulate a human-like interaction, it is useful to include also nonverbal language, such as body language, which is possible through movements and gestures that Pepper can accomplish but also with colours that it can use thanks to its coloured led on its body. Pepper is also equipped with some sensors to get some information like for example the distance of nearby objects and people. This is perfect to adapt the robot's behaviour and to respect prossemic rules adapting to the users' personalities which is fundamental to gain their trust and engagement.

The proposed solution gives to Pepper the assignment to provide to the social routine of seniors: it will be a playmate, an entertainer, a confidant, and the intermediary to maintain or establish new relationships. Pepper could entertain users with dance and singing exhibitions if they ask to, with some minigames (memory, akinator, crossword puzzle), being a storyteller or telling jokes. Pepper will be designated to accomplish repetitive and boring tasks to provide to the lack of the staff. Some examples could be to show the way to seniors inside the structure trying to detect and point obstacles on the way to safeguard the users from falling. Then Pepper has the assignment to answer recurring questions such as questions about weather, calendar information and update about the latest news. However, Pepper cannot completely replace a human caregiver, for this reason if it cannot help efficiently a user a caregiver will be called by the SAR to better answer the user's problem.

Pepper will also be a coach/ therapist and weekly seniors could attend some different activities managed by the SAR. The proposed activities aim to stimulate physical and brain capabilities but also to amuse them with therapeutic purposes, sometimes working on the collaboration between participants and some others playing creating different teams to stimulate competition. Actually, some of the proposed activities are classical games like bingo, exercises to train cognitive abilities and physical exercises. Specifically, some imagined activities are based on the already mentioned recollection power of the music which can help in brain training but also to better engage users. Actually, all these activities must be customised according to seniors' abilities, preferences and personalities, and in accordance with their preferential choices [52].

In the same way, physical exercises are customised to the senior target. These activities are thought to be performed in groups to allow and facilitate the maintenance and establishment of new relationships between the participants.

As already mentioned, the proposed activities do not require the constant participation of a caregiver. However, it means that users with severe mental disease cannot participate in this specific case. On the other hand, if a caregiver would be willing to take part constantly in the activity also seniors with cognitive impairments would be allowed.

Anyway, Pepper could also be part of other custom-made activities for that different kind of user. An example could be adapting some exercise to the human-robot interaction system, such as the "Time Slips" an activity created by Anne Basting [44] making robot and caregiver cooperate. Actually, the activity consists in making patients with cognitive impairments develop a story, taking inspirations from photos, images and input given by Pepper with the help of a caregiver.

In the end, thanks to Pepper's tablet, nursing home guests can video call relatives and friends to keep in touch with them and not feel abandoned anymore. At the same time, Pepper's tablet also allows the implementation of telemedicine when needed.

### 6.2.3  Chair exercises for elderly application

Taking inspiration from the above-mentioned German research CIT CARROS, a physical activity for seniors has been developed with the aim to train seniors weekly. In particular, Pepper coaches its users following some specific exercises called "chair exercises" which can be performed, as a matter of fact, sitting on a chair to be less strenuous and custom-made for seniors' ability. The chair exercises are imagined to be performed in groups not only because of the establishment of new friendships purposes. Executing this kind of activity together can bring more engagement and fun during the activity. Actually, research proved that seniors learn to rely on each other, comparing their exercises with the other participants to better understand what the right movement is to perform for a better final result.

As already said, the activity can be executed without the constant presence of a caregiver, unless patients with cognitive impairments take part of it. Even though the presence of a caregiver could help in any case in the accomplishment of the activity, he can also encourage users to be engaged and to participate in the interaction reassuring them. Actually, at the beginning of the activity users could be a little bit intimidated or shy, so that the presence of a caregiver at this point of the interaction could be useful to support them, also helping Pepper in its purpose.

The activity aims to accompany the group along the entire session: at first all along the introductory part of the activity, where Pepper explains rules and what kind of activity they are going to do, then going through the warm-up session and finally executing the real training activity. All along the training Pepper has the assignment to safeguard seniors' health and to monitor their moods. The latter is useful to register seniors' feedback about the exercises, the obtained engagement and, at least, to answer in a proper way. As a matter of fact, Pepper should encourage participants in the activity, monitoring especially bad moods. If bad moods are detected Pepper should comfort seniors and incite them assuming the role of a motivational speaker. At the same time, throughout the exercises, Pepper must remind seniors not to push themselves too much if the movements are too hard compared to their abilities and movement possibilities. The reminder must be repeated more than once all along the session because Pepper cannot verify with its cameras if someone hurts himself. Therefore, Pepper must normalize this kind of inability, supporting seniors, giving the suggestion of stopping the movement and starting again the session with the following one, even trying not to be boring.

As a matter of fact, the presence of a caregiver could be helpful also in monitoring seniors' state of health but, if it would not be possible, a trick which could be used is to elect a group leader among seniors. Opting for the election of a leader seems to be also a good solution to make the oral interaction simpler and clearer, so that Pepper could better understand if any problems are there, to manage them in the best possible way. Moreover, giving to the user a role who has responsibilities creates more engagement and makes the user feel a sense of control of the interaction, which is fundamental for the user who does not feel overwhelmed by the system. Indeed, the group leader will be called by his name every time the robot needs to. In this way a relationship made of trust will be better established between the group leader and Pepper but also between the group leader and other users who have to rely on him. Indeed, during the exercise session, Pepper will ask the group leader if it can go on with the activity, if it has to repeat something which is not clear or if they need to stop the session because someone needs the caregiver's assistance. Nevertheless, to stop the interaction only if necessary it is essential to introduce a control system to stop it at the right moment: when the group leader asks to stop the session, Pepper is required to ask if it is needed for real, that's because the robot will call a caregiver to give them support. In this way it will be possible to repair the interaction and return to the previous activity. As already mentioned, Pepper is also required to repeat instructions if the group asks for it. That's because every exercise is before verbally explained and then shown with Pepper movements, so that the explanation could be as clear as possible.

Finally, at the end of the training, Pepper proposes a cool-down session where the robot suggests some deep breaths to relax muscles, then it thanks users for being its playmates and asks for feedback keeping track of their moods before closing the interaction.

# 7 Conclusions

Over the past few years, Human- Robot Interaction applied in social robotics and seniors therapy benefit from the huge progresses made by the AI. In addition, world aging problems, as explained above, helped with an even wider robotics usage in western countries in elderly care as an efficient solution for caregiver shortage in recent years. Nevertheless, AI still has some limits, such as entry barriers, sarcasm and falsehood detection, making some behavioural models' comprehension still difficult for an informatic system.

Most socially assistive robotics experimentations with seniors comprehends short-term and medium-term tests, often lasting a few weeks or months, making a clear engagement evaluation hard to define in the long-term.

The seniors often lack experience and knowledge about the socially assistive robots, often unfamiliar with technology and with stereotypical convictions about social robotics, which make it more difficult for elderly to accept SARs.

Moreover, potential users are not always willing to pay high initial costs for a SAR, representing a burdensome entry barrier for this kind of assistive systems.
However, the users taken into consideration in the experimentations related to studies, demonstrated themselves prone to utilisation of SARs in assistive context, enjoying their benefits, foreshadowing a future in which the socially assistive robotics could result widespread in the life of people and in care facilities.